# Water Care: Water Surface Cleaning Bot and Water Body Surveillance System

Harsh Sankar Naicker[1], Yash Srivastava[2], Akshara Pramod[3], Niket Paresh Ganatra[4], Deepakshi Sood[5], Saumya Singh[6], Velmathi Guruviah[7]
School of Electronics Engineering, Vellore Institute of Technology Chennai Campus, Chennai, India - 632014
Email: 1harshunaicker@gmail.com, 2yash33srivastava@gmail.com, 3aksharuhi@gmail.com, 4 niketganatra06@gmail.com , 5deepakshisud@gmail.com, 6saumyasingh3103@gmail.com, 7velmathi.g@vit.ac.in

*Abstract* - **Whenever a person hears about pollution, more often than not, the first thought that comes to their mind is air pollution. One of the most under-mentioned and under-discussed pollution globally is that caused by the non-biodegradable waste in our water bodies. In the case of India, there is a lot of plastic waste on the surface of rivers and lakes. The Ganga river is one of the 10 rivers which account for 90% of the plastic that ends up in the sea (Source: Sky News) and there are major cases of local 'naalas' and lakes being contaminated due to this waste. This limits the source of clean water which leads to major depletion in water sources. From 2001 to 2012, in the city of Hyderabad, 3245 hectares of lakes dissipated. The water recedes by nine feet a year on average in southern New Delhi. Thus, cleaning of these local water bodies and rivers is of utmost importance. Our aim is to develop a water surface cleaning bot that is deployed across the shore. The bot will detect garbage patches on its way and collect the garbage thus making the water bodies clean. This solution employs a surveillance mechanism in order to alert the authorities in case anyone is found polluting the water bodies. A more sustainable system by using solar energy to power the system has been developed. Computer vision algorithms are used for detecting trash on the surface of the water. This trash is collected by the bot and is disposed of at a designated location. In addition to cleaning the water bodies, preventive measures have been also implemented with the help of a virtual fencing algorithm that alerts the authorities if anyone tries to pollute the water premises. A web application and a mobile app is deployed to keep a check on the bot's movement and shore surveillance respectively. This complete solution involves both preventive and curative measures that are required for water care.**

## I INTRODUCTION

1. Objectives and Goals

The amount of waste keeps on increasing worldwide in water bodies. Food containers and packaging are the largest components of the solid waste stream (80 million tons or 31.7%). This garbage (80%) comes from trash and garbage in the municipal sewer system, that is, from land sources. These plastic wastes affect the marine ecosystem. Hence it is necessary to clean the waste in the ocean and make the water pollution free. However, effective cleaning mechanisms have not been deployed in India to take care of the situation. Hence, we have decided to come up with an invention to collect the waste in the water bodies and make the water free from pollution.

2. Applications

The first application is that of garbage collection from water bodies by the bot. The second application is that of prevention of littering of water bodies by the help of virtual fencing on the shore by implementing image processing on the CCTV feed.

3. Features

The entire solution provided is cost effective when compared to existing solutions. The bot that is being created is powered using renewable energy which makes it Eco friendly and sustainable. The solution can be used in any water body. Implementation of image processing to collect waste thrown by humans in the waterbody as well as prevention of such incidents is included.

## II LITERATURE REVIEW

Namami Gange Programme [1] was launched by the Union Government, and includes sewage treatment, industry effluent monitoring. The objective was conservation, effective abatement of pollution, and rejuvenation of the national river Ganga. But it is not specifically focused on automated or controlled devices for river cleaning. A. Sinha et al. [2] proposed an unmanned ship Ro-Boat, which can detect, collect and remove debris, chemical sewage present on the surface autonomously. However, there is no provision for the security of bots in water. Robot SEARCh [3] is used to clean water bodies(specifically rivers), consisting of a raft-like structure made of PVC pipes and rubber. It has a conveyor belt that collects the garbage, but no power source mentioned, and collects it on the raft itself, weight constraints not mentioned. It is cheap, uses recycled materials to build the bot, but there is no controllability of the bot once it is put to action. Trash Skimmers [4] are equipment that helps remove floating waste from the river surface at popular ghats. It is a large-scale project for cleaning massive amounts of garbage. It does not segregate different types of trash, large and so cannot be equipped on every river or pond. R. Raghavi et al. [5] proposed a remote-controlled cleaning bot along with a pH sensor that determines the solubility and biological activity of the chemical constituents of water. The proposed work focuses on monitoring the water quality and also collects the garbage waste that is floating on the water surface. E Rahmawati et al [6] describe the design of a robot for cleaning rubbish floating on the water surface. They developed a pontoon-shaped hull that fulfills all the hydrostatic and structural criteria of the boat. The hull can bear maximum trash of up to 16 kg. Siddhanna et al [7] proposed a robotic arm that can detect, pick, and place garbage from water bodies and thereby clean the water bodies. The proposed system is embedded with sensors for detecting obstacles and their respective distances from the boat and identifying whether the organisms are living or nonliving. Xiali Li et al [8] proposed a modified YOLO v3-based garbage detection method, allowing real-time and high-precision object detection in dynamic aquatic environments which enhanced the robot's performance. Lucia Maddalena et al [9] proposed an approach based on self-organization via artificial neural networks. Their work was useful for video surveillance systems like applications. Harsh Panwar et al [10] proposed a dataset called AquaTrash which is derived from the TACO data set. They then proposed a deep learning-based object detection model called AquaVision which classifies different pollutants that can be found floating on water bodies. Hsing-Cheng Chang et al [11] proposed a multi-function unmanned surface vehicle which has fourfold features - autonomous obstacle avoidance and navigation; water quality monitoring, sampling, and positioning; water surface detection and cleaning; and remote navigation control and real-time information display. M. N. Mohammed et al [12] proposed a design of a rubbish collection system. Arduino is used as the microcontroller for driving DC motors which draw power from solar energy and features ultrasonic sensors for distance measurement to the nearest obstacle. Xiahong Gao and Xijin Fu [13] proposed a miniature water surface garbage cleaning bot. STC12C5A60S2 controller is used along with WiFi for wireless control. A mobile app is developed to send instructions to the controller. A lightweight prototype is developed, but is restricted to a range of 20 meters and is not waterproofed. Akash et al. [14] proposed a machine in the paper that lifts debris from water surfaces and disposes of it within the tray that is inbuilt in the machine using a conveyor. The machine is manned using an RF transmitter and receiver that manage the remote remotely. It is easy to operate and flexible and uses a renewable source of energy. However, there is limited capacity to the tray that collects the garbage and only the waste that is floating on the surface of the water can be collected. S. S. Hari et al. [15] have proposed a bot that gathers the trash from lakes with the assistance of a transport line and stores it in the trash bin. By setting up an association with the Node-MCU board by utilizing Wi-Fi convention, the bot handles the direction of propellers which assists the robot with exploring the water bodies. Having only one type of connection limits the use of bots in larger areas as the range cannot be manually extended from the point of operation. Jayashree et al. [16] aimed to design, develop and demonstrate a new open hardware and software technological platform to improve the monitoring of water bodies and assess their healthiness in real-time. The bots focus on increasing water health using the system of bots that work together interconnected using a radio system. The robots are connected to a land station that needs constant monitoring which compromises autonomy. Moreover, since the bots are manned underwater as well, if one of the bots goes missing, it is difficult to locate and remove the bot without physical labor. Soumya et al. [17] talk about "Pond Cleaning Robot", which involves removing waste debris from the water surface and disposing it safely. A bluetooth module and DC motors are interfaced with the microcontroller. LEDs are used to indicate the direction of the robot. Firdaus, D., Priambodo, B. & Jumaryadi, Y. [18] explained the process of using push notifications in Kotlin driven Android projects using Firebase, enabling the surveillance system created to send alert notifications to an android application when there is a possibility of someone polluting the water body. Ovidiu, Stan and Liviu Miclea [19] created a system similar to what we aim to implement. They have used a live camera feed from a Pi cam which has been hosted on port 8000 which can be accessed by the application using a RESTful API service.

III HARDWARE AND SOFTWARE COMPONENTS

HARDWARE:
Raspberry Pi 3B: The Raspberry Pi 3B is a small but very capable single board computer (SBC). It comprises a lot of practical features such as a CSI port for connecting a Picamera, 4 USB-2 ports, ethernet and WLAN connectivity, BLE. It uses a Broadcom BCM2837 Quad Core 1.2GHz CPU which offers upto 1GB RAM. The micro SD card slot can be used for data storage. 40 pin GPIO is another very practical feature of the Rpi which facilitates interfacing sensors and other peripherals. The Raspberry Pi can run many available OS, the most popular one is the Raspberry Pi OS, although other distributions of Linux may also be used.
Arduino Uno: The Arduino Uno is a very popular microcontroller board that is widely used in embedded systems and IoT applications. At its core is the ATmega328P microcontroller. The greatest virtue of the arduino is it's robust digital and analog pins which are very easy to use and interface with motor drivers, sensors, relays etc. 6 of the 14 digital pins support pulse width modulation, which enables us to set variable speeds for motors and such. The arduino can be interfaced directly with the Raspberry Pi using the USB port, from which it can both draw power as well as communicate with the Rpi using the pyserial library.

Solar Panel rated at 18V- 55 watt, composed out of Polycrystalline Silicon. It produces a max output of 7.2 V and 3 W.  Size: 230 x 140 mm.

4 DC motors. Two motors will be placed on either side of the chassis for propulsion. Two motors will be used to control the conveyor belt.

Polyvinyl conveyor belt. Controlled by the arduino using a relay circuit. Collects all floating garbage. Belt Capacity: 14 kg. Belt Speed: 5-35 ft/min Belt Width: 10*60 mm. Belt Length: 457 mm. Roller Diameter: 10mm

Pi-camera V2: 8MP resolution, interfaces with the Rpi using CSI port. Can be utilised for live video streaming as well as snapshots.

Battery: Maintenance free sealed Lead-acid battery

GPS: Ublox M8N GPS module

Pixhawk 2.4.8: The Pixhawk 2.4.8 is a well known flight controller and is very established in UAVs. Its processor is a 32-bit ARM Cortex M4 with 256KB RAM at 168 MHz and 2MB of flash memory. It has an inbuilt accelerometer, barometer, magnetometer and gyroscope which enables it to have a sense of orientation, velocity, and altitude at any instant of time. It has SPI, I2C, CAN protocol enabled ports, along with 2 telemetry ports along with SBUS and PPM sum signal functionalities. We are using the pixhawk along with the boat firmware which is available with ardupilot.

SOFTWARE:

Software has three main modules: (i)Surveillance & Detection - using Computer Vision concepts, (ii)Web Application - for viewing the livestream from the bot and, (iii)Mobile Application - for alerts from the Surveillance System.

OpenCV Python - OpenCV is an open source library which deals with image processing and other computer vision related tasks. It offers a plethora of functions and methods which facilitate many image processing tasks. OpenCV is majorly used with C++ and python, and in our case it is the latter.

PostgreSQL - The detections made using the computer vision algorithm will be committed to the cloud PostgreSQL database. This will allow the web application and the mobile application to fetch data from the database and send alert notifications to the concerned authorities as and when required.

Flask Framework in Python - The Flask Framework will be used to create the APIs that the web application and the mobile application will use to get data back from the PostgreSQL database and also to get the video feed from the Picam attached to the raspberry pi for surveillance.

Node.js - Node.js is single-threaded which makes it ideal for creating event-driven servers. Traditional websites and API services It's used for traditional web sites and back-end API services along with real-time and push-based architectures for

Express.js - Express is a Node.js webapp framework which is flexible and can be used for easy development of web and mobile applications using its various powerful set of features. All the fundamental features are provided by Express without obscuring Node.js features. Additionally, it also forms the basis for various other frameworks.

HTML,CSS,JS - HTML will be  used to define the structural information of the webpage. HTML,CSS,JS are indispensable and important elements of websites. CSS is mainly responsible for the look and feel of the website, commonly referred to as the front-end and JavaScript (JS) is used for providing programming functionality.

EJS - EJS is a language which lets users generate HTML code using JavaScript. Its features include fast compilation and rendering, support for both, server JS and browser, etc.

Firebase(Authentication and Authorization) - We will be using Firebase as Backend as a Service (BaaS). This means that the server side processing ranging from user authentication, real time database, push notifications, etc. will all be handled by Firebase. On the client side, which is the User Interface (UI) that a user interacts with, the client side logic will be written. All of the heavy processing will be handled by Firebase.

Heroku (Hosting) - Heroku is a Platform as a Service (PaaS) that provides free cloud hosting services for a variety of applications made using different technology stacks. We will be using Heroku to deploy our web application for anyone to use online. The PostgreSQL that will be used in our surveillance system will be acquired using the add-on feature in Heroku itself.

Android Studio - Android Studio will be used for compiling the SDK of the mobile application. This uses Java in the backend for compiling the necessary binaries required for running a mobile app on an Android device. Furthermore, provisions for using a virtual android device are provided for app testing which speeds up the development process.

Google Firebase - We will be using Firebase as Backend as a Service (BaaS). This means that the server side processing ranging from user authentication, real time database, push notifications, etc. will all be handled by Firebase. All of the heavy processing will be handled by Firebase.

XML - XML is used for the UI definitions in the mobile app. This means that the layout of the app is set up using XML.

Kotlin - Kotlin is developed by JetBrains which is a statically typed language. It is open-source which uses the LLVM compiler technology to compile Kotlin sources. They are compiled into stand-alone binaries for different operating systems and CPU architectures like Mac, Windows, Linux, iOS,  and Webassembly.

Ardupilot - Ardupilot is an open source software project that deals with developing and providing top-of-the line autopilot systems. It has firmware for a plethora of vehicles which include multirotors (quadcopters, hexacopters etc.), rovers, vertical take-off and land (VTOL) planes, and boats. Ardupilot's open source nature makes it forever improving with a lot of community support and therefore it is adopted by many organizations and universities for research and development.

SITL - Software in the Loop simulators are used to test the performance of vehicles in a virtual environment without making use of any physical hardware. We are using Mission Planner's simulation features which have SITL which contains ardupilot's core APIs.

Mission Planner - The Mission Planner is a powerful Ground Control Station which is mainly compatible with

Windows OS. Using mission planner we can directly load firmwares sourced from ardupilot, and run SITL simulations. It can also be used to set waypoints and create mission logs, tune vehicle parameters and modify PID values as well. It supports all ardupilot based vehicles.

## IV METHODOLOGY

HARDWARE:
The hardware implementations comprises i) Assembly of the Power system for sustainable power supply ii) Integration of the power system with the bot and embedding a conveyor belt for trash collection and disposal. iii) Navigation

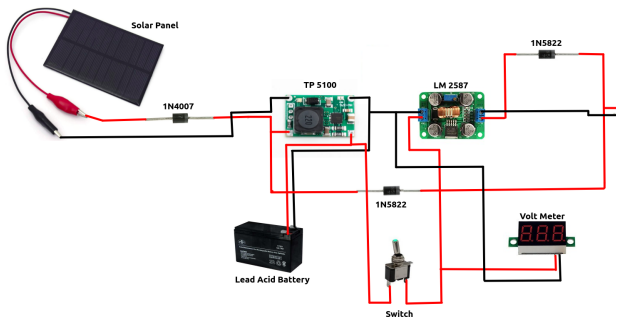

Fig 1 - Circuit diagram for the power system

Power System - In normal conditions, power from the mains is drawn by the solar panel to charge the lead acid battery and to provide power to the motor. When the solar power fails, the stored energy in the battery is used to power up the motor. The voltmeter display is used to display the battery voltage level. The two diodes 1N5822 are used to block the reverse current flow. The positive terminal input DC jack is connected to the positive terminal of the output DC jack through a Schottky diode ( 1N5822 ). The 12V input power from the solar panel is connected to the input terminal of the TP5100 module through a 5.5mm DC Jack. The output terminal of the TP5100 charging module is connected to the battery. The battery positive terminal is connected to the boost converter LM2587 IN+ terminal through a switch and the negative terminal is directly connected to boost converter IN- terminal. The boost converter LM2587 Out+ terminal is connected to the positive terminal of the output DC Jack through a Schottky diode ( 1N5822 ) and the Out- terminal is connected directly to the negative terminal of the DC jack. The voltmeter positive terminal is connected to boost converter IN+ and the negative terminal is connected to IN-.

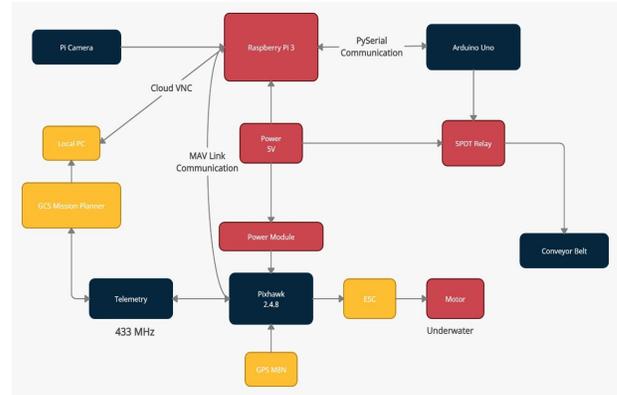

Fig 2 - Block diagram of the components

The Raspberry Pi draws 5V input from the power system's output and performs the following tasks: (i) Run image processing algorithm on picamera's video input for trash detection and generate control signals for bot's motion and the conveyor belt's operation. The motion is handled by the Pixhawk, hence the control signal is communicated to it via the MAVLink protocol. The control signals for the conveyor belt are communicated to the Arduino via the PySerial library. (ii) Broadcast the live video stream to the offshore local PC.. This is done by using the cloud services offered by RealVNC. The arduino gets the control signal from the Raspberry Pi for activating the conveyor belt, and it does so by combining the logic with 5V power from the power system which is fed in using an SPDT relay. The navigation is handled by the Pixhawk board, which is a flight controller and contains an inbuilt accelerometer for speed measurement, gyrometer for roll, pitch measurements and an IMU for orientation. The pixhawk is interfaced with a GPS module using which it follows a crow's-flight path to the next waypoint. The kinematic parameters of the bot can be tracked in real time via the telemetry connection to a GCS enabled PC.

SOFTWARE:
For the surveillance mechanism, a camera will be enabled at the shore to detect humans that try to pollute the water bodies using image processing algorithms. The image processing algorithms are used for human detection. Based on the detection, an alert will be sent to the authority via app to take appropriate measures. The image processing algorithm is used to detect human action on the coast using opencv and deep learning models to classify the action (polluting water bodies/not polluting water bodies). For accessing the real time video feed generated by the cameras (both the one mounted on the bot and the one used for offshore surveillance) on the app, we use the custom TCP tunnel service offered by Remote.it. Remote.it establishes a tunnel connection between TCP port 8081 of the Raspberry Pi (which runs motion) and the end device which allows the video stream to be accessed remotely.

When it comes to the alert system, we need to send an alert notification to the mobile application whenever someone crosses the virtual fence, i.e., when the count increases. This is done by using Google Firebase. The python script which is used for the purpose of virtual fencing and detection of crossing of the virtual fence is connected to the same Firebase project as the mobile app. So, whenever the count increases, a push notification is sent to the device which has the application using Google Firebase and upon receiving the notification, the authority who has the access to the application and hence the live feed of the bank of the water body, can immediately check who is the perpetrator. This app would be present only with the authorities like village Sarpanch, or selected officials whose job is to monitor the banks of the river for possible littering. The live feed of the camera at the bank of the water body has been accessed using a simple Browser Intent [15] in Kotlin which is run when the button to view the stream is clicked using the setOnClickListener function [16]. A web application will also be included for the same build with Express.js and HTML, CSS and Javascript. It will be connected to the same Google Firebase app for the purpose of authentication, hence providing a uniform system. The live stream from the bot can be viewed on the web application along with periodic alerts and notifications. So, the bot awareness will be increased.

## V RESULTS

A two pronged solution that is both preventive and reactive has been proposed. The end-to-end ecosystem that connects the bot deployed on the water on one end, to the cameras on the shores of water bodies to the mobile devices of local authorities gives a complete solution for water care. The proposed approach consists of a Mobilenet-SSD based computer vision algorithm that detects garbage floating on water bodies. The concept of virtual fencing introduced is unique which correctly tracks moving bodies and alerts when the virtual fence set in the frame is crossed. Lastly, the prompt notification system proposed for the local authorities is a reactive measure which can be clubbed with on site logistics to initiate prompt actions on the shores where people may come to visit. Lastly, the solar power system makes the proposed solution sustainable and in addition to the autonomous capability of the bot, truly removes human intervention.

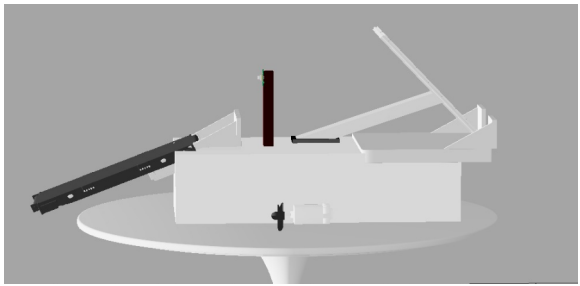

Fig 3 - Side view

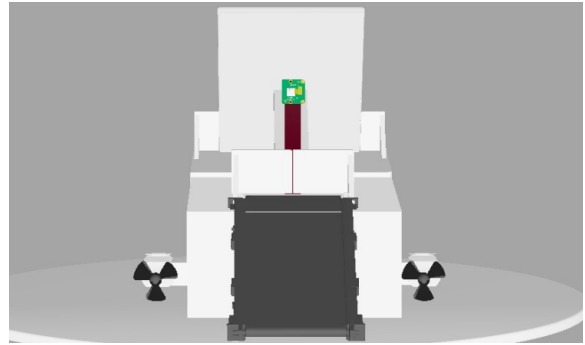

Fig 4 - Front view

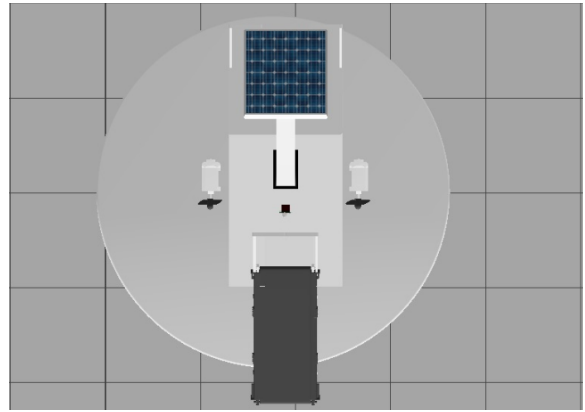

Fig 5 - Top view

Figure 3, 4, 5 illustrate the different views of the 3D model of the bot created using Autodesk Maya with Arnold render.

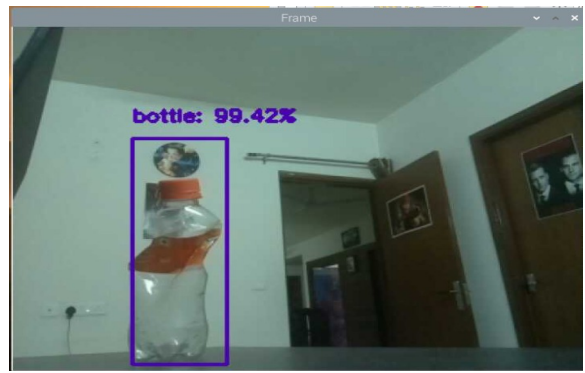

Fig 6 - output of the trash detection algorithm

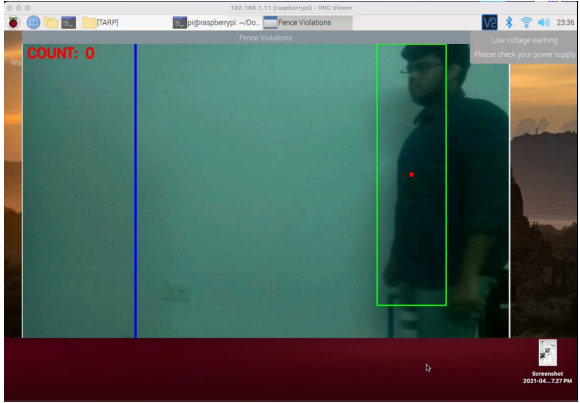

Fig 7 - before a person crosses the virtual fence, the count is zero

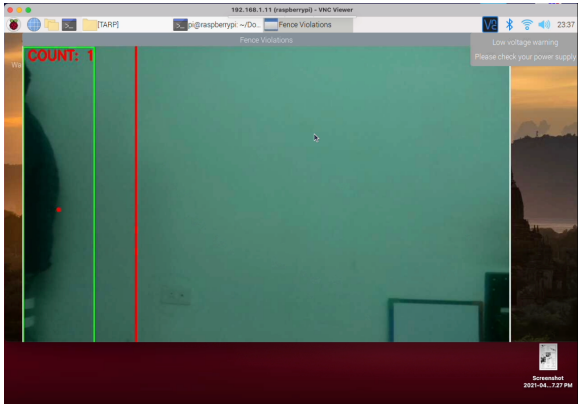

Fig 8 - after a person crosses the virtual fence, the count is incremented to one

Fig 7 and 8 show the output of the virtual fencing algorithm.

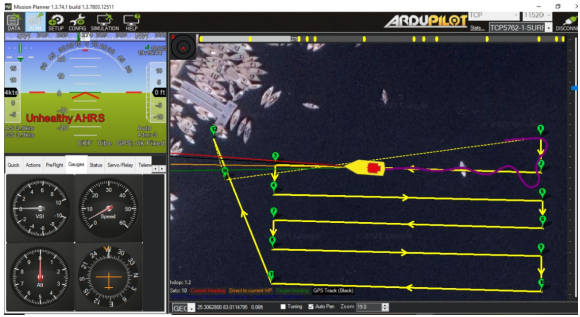

Fig 9 - Autonomous navigation of the robot in SITL

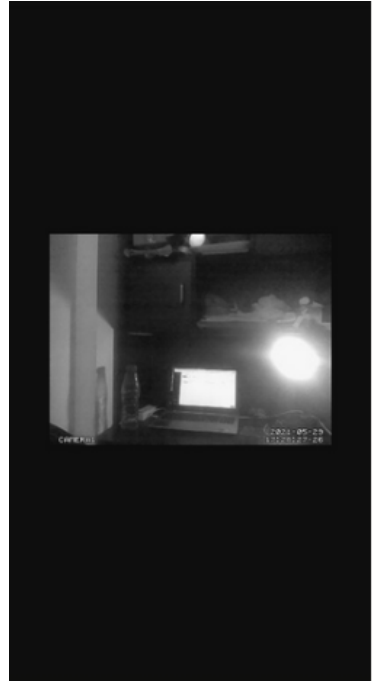

Fig 10 - Live feed that would show the bank of the water body and can be accessed via the mobile app.

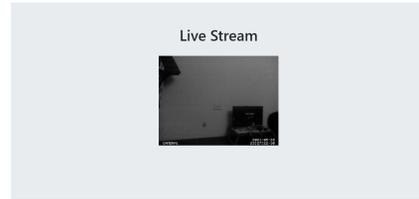

Fig 11 - Live feed of the camera attached on the bot that can be accessed using the website.

## VII CONCLUSION

A complete and sustainable solar powered solution for preserving water bodies is offered, which consists of a water-surface cleaning robot along with onshore surveillance. Usage of open-source tech stacks ensures that the code base is cutting edge and low cost. The proposed solution is both proactive and reactive. The algorithm for detecting garbage needs to be trained rigorously for improved speed and accuracy in detection of garbage in dynamic scenes captured in image frames. Size of the robot must be vertically scaled for real water bodies and a greater cap for garbage collection. A fleet of robots may be required to be utilized depending upon the area of the water body to be covered and the average output that a single robot exhibits upon on-site testing. Power system of the bot needs to be tested after deployment and modifications can be made to maximise efficiency. Finally, the optimised solution can be directly put into use for preserving water bodies.

## VII REFERENCES

1. National Mission for Clean Ganga(NMCG), G., 2021. नमामि गंगे. [online] National Mission for Clean Ganga(NMCG), Ministry of Jal Shakti, Department of Water Resources, River Development & Ganga Rejuvenation, Government of India. Available at: https://nmcg.nic.in/NamamiGanga.aspx
2. A. Sinha, P. Bhardwaj, B. Vaibhav, and N. Mohommad, "Research and development of Ro-boat: an autonomous river cleaning robot," NASA/ADS. [Online]. Available: https://ui.adsabs.harvard.edu/abs/2013SPIE.9025E..0QS/abstract
3. Bhatkhande, Ankita. "Mumbai Students Build Robot to Help Clean Surface Water." DNA India, 23 Mar. 2017, https://www.dnaindia.com/india/report-mumbai-students-build-robot-to-help-clean-surface-water-2364246.
4. "Trash skimming — Cleantec Infra," Cleantecinfra.com. [Online].Available: https://www.cleantecinfra.com/trash-skimming.
5. "Water Surface Cleaning Robot", R. Raghavi1, K . Varshini2, L. Kemba Devi3 [Online] Available: http://www.ijareeie.com/upload/2019/march/42_NCIREST106.pdf
6. E Rahmawati et al 2019 J. Phys.: Conf. Ser. 1417 012006
7. Siddhanna Janai , H N Supreetha , Bhoomika S , Yogithashree R P, Pallavi M, 2020, Swachh Hasth-A Water Cleaning Robot, INTERNATIONAL JOURNAL OF ENGINEERING RESEARCH & TECHNOLOGY (IJERT) Volume 09, Issue 07 (July 2020),
8. Li, Xiali & Tian, Manjun & Shihan, Kong & Wu, Licheng & Yu, Junzhi. (2020). A modified YOLOv3 detection method for vision-based water surface garbage capture robots. International Journal of Advanced Robotic Systems. 17. 172988142093271. 10.1177/1729881420932715.
9. L. Maddalena and A. Petrosino, "A Self-Organizing Approach to Background Subtraction for Visual Surveillance Applications," in IEEE Transactions on Image Processing, vol. 17, no. 7, pp. 1168-1177, July 2008, doi: 10.1109/TIP.2008.924285.
10. Panwar, Harsh & Gupta, Pradeep & Siddiqui, Mohammad Khubeb & Morales-Menendez, Ruben & Bhardwaj, Prakhar & Sharma, Sudhansh & Sarker, Iqbal. (2020). AquaVision: Automating the detection of waste in water bodies using deep transfer learning. Case Studies in Chemical and Environmental Engineering. 2. 100026. 10.1016/j.cscee.2020.100026.
11. Chang, Hsing-Cheng & Hsu, Yu-Liang & Hung, San-Shan & Ou, Guan-Ru & Wu, Jia-Ron & Hsu, Chuan. (2021). Autonomous Water Quality Monitoring and Water Surface Cleaning for Unmanned Surface Vehicle. Sensors. 21. 1102. 10.3390/s21041102.
12. M. N. Mohammed, S. Al-Zubaidi, S. H. Kamarul Bahrain, M. Zaenudin and M. I. Abdullah, "Design and Development of River Cleaning Robot Using IoT Technology," 2020 16th IEEE International Colloquium on Signal Processing & Its Applications (CSPA), 2020, pp. 84-87, doi: 10.1109/CSPA48992.2020.9068718.
13. X. Gao and X. Fu, "Miniature Water Surface Garbage Cleaning Robot," 2020 International Conference on Computer Engineering and Application (ICCEA), 2020, pp. 806-810, doi: 10.1109/ICCEA50009.2020.00176.
14. Akash Shahu, 2021, Remote Controlled Unmanned River Cleaning Bot, INTERNATIONAL JOURNAL OF ENGINEERING RESEARCH & TECHNOLOGY (IJERT) Volume 10, Issue 03 (March 2021),
15. S. S. Hari, R. Rahul, H. Umesh Prabhu, and V. Balasubramanian, "Android Application Controlled Water Trash Bot Using Internet Of Things," 2021 7th International Conference on Electrical Energy Systems (ICEES), 2021, pp. 538-542, doi: 10.1109/ICEES51510.2021.9383698.
16. Jayashree dadmal, Saloni Bardei, Namira Khan, R. M. I. S. S. R. W. "RIVER TRASH COLLECTION BOT". Design Engineering, July 2021, pp. 1149-58, http://www.thedesignengineering.com/index.php/DE/article/view/2556.
17. Soumya, H. M., & Preeti, B. G. (2018). Pond Cleaning Robot. International Research Journal Of Engineering And Technology (Irjet).
18. Firdaus, D., Priambodo, B. & Jumaryadi, Y. (2019). Implementation of Push Notification for Business Incubator. International Association of Online Engineering.
19. Ovidiu, Stan, and Liviu Miclea. "Remotely Operated Robot with Live Camera Feed." International Journal of Modeling and Optimization 9.1 (2019).